\documentclass[a4paper]{cas-dc}

\usepackage[numbers]{natbib}

\usepackage{cleveref}
\usepackage{booktabs} 
\usepackage{longtable}
\usepackage{graphicx}

\def\tsc#1{\csdef{#1}{\textsc{\lowercase{#1}}\xspace}}
\tsc{WGM}
\tsc{QE}
\tsc{EP}
\tsc{PMS}
\tsc{BEC}
\tsc{DE}

\begin{document}
\let\WriteBookmarks\relax
\def\floatpagepagefraction{1}
\def\textpagefraction{.001}


\shortauthors{Feng Shuang et~al.}

\title [mode = title]{3D Hand Reconstruction via Aggregating Intra and Inter Graphs Guided by Prior Knowledge for Hand-Object Interaction Scenario}                      



%
\author[1]{Feng Shuang}[style=chinese,
                        orcid=0000-0002-4733-4732]



\ead{fshuang@gxu.edu.cn}



\affiliation[1]{organization={Guangxi Key Laboratory of Intelligent Control and Maintenance of Power Equipment, School of Electrical Engineering, Guangxi University},
    addressline={100 Daxuedong Rd.}, 
    city={Nanning},
    postcode={530004}, 
    state={Guangxi},
    country={China}}

\author[1]{Wenbo He}[style=chinese]

\author[1]{Shaodong Li}[style=chinese
   ]
\cormark[1]
\ead{lishaodongyx@126.com}





\cortext[cor1]{Corresponding author}




\begin{abstract}
Recently, 3D hand reconstruction has gained more attention in human-computer cooperation, especially for hand-object interaction scenario. However, it still remains huge challenge due to severe hand-occlusion caused by interaction, which contain the balance of accuracy and physical plausibility, highly nonlinear mapping of model parameters and occlusion feature enhancement. To overcome these issues, we propose a 3D hand reconstruction network combining the benefits of model-based and model-free approaches to balance accuracy and physical plausibility for hand-object interaction scenario. Firstly, we present a novel MANO pose parameters regression module from 2D joints directly, which avoids the process of highly nonlinear mapping from abstract image feature and no longer depends on accurate 3D joints. Moreover, we further propose a vertex-joint mutual graph-attention model guided by MANO to jointly refine hand meshes and joints, which model the dependencies of vertex-vertex and joint-joint and capture the correlation of vertex-joint for aggregating intra-graph and inter-graph node features respectively. The experimental results demonstrate that our method achieves a competitive performance on recently benchmark datasets HO3DV2 and Dex-YCB, and outperforms all only model-base approaches and model-free approaches.
\end{abstract}



\maketitle
\section{Introduction}
3D hand reconstruction has been widely used in human-machine interaction \cite{conci2007natural}\cite{yin2021wearable}, virtual and augmented reality \cite{han2020megatrack}\cite{jung2016body}\cite{mueller2019real}\cite{wang2020rgb2hands}, sign language recognition and translation \cite{liang2020multi}, and has made great progress in recent years. But it is still challenging to robust to occlusion, particularly when the hand is severely occluded under hand-object interaction scenario.

The current approaches of 3D hand reconstruction focus on recovering highly accurate and plausible 3D hand meshes. Those approaches can be generally categorized into model-based and model-free approaches.  Model-based approaches \cite{boukhayma20193d}\cite{chen2021model}\cite{ge20193d}\cite{park2022handoccnet} refer to utilize a parametric model with strong prior knowledge of hand and regress parametric representation in terms of shape and pose. Among the approaches in recent years, MANO \cite{romero2022embodied} as statistical parametric model is the most widely used and shows great advantages in reconstructing physical plausible hand model when facing the challenges of object-occlusion, self-occlusion and low resolution. However, the regression of accurate MANO parameters from RGB image is a highly nonlinear mapping process, and will be limited since the MANO parameter model is constructed using limited hand samples. In contrast, Model-free approaches \cite{choi2020pose2mesh}\cite{hampali2022keypoint}\cite{kolotouros2019convolutional}\cite{lin2021end} directly regresses to the coordinates of 3D hand mesh vertices and joints, and no longer employ predefined parameterized model. Model-free approaches can recover highly accurate 3D hand meshes and joints because of measuring an average joint or vertex position error directly. However, model-free approaches can not robust to meet physical plausibility because of without prior information of hand shape and pose when facing challenging environment variations.

Therefore, it is necessary to combine the benefits of model-based and model-free approaches. Most recently, Jiang et al. \cite{jiang2023probabilistic} propose a probabilistic model that estimate the prior probability distribution of joints and vertices with incorporating a model-based network as a prior-net. Yu et al. \cite{yu2023overcoming} leverage direct mesh fitting for alignment accuracy and guidance from MANO for plausibility. Motivated by the above observations, our ultimate goal is to reconstruct both highly accurate and physical plausible hand model with the benefits of model-based and model-free methods for challenging hand-occlusion. In this work, we propose a 3D hand reconstruction network cascading MANO and Model-free method that consists of initial stage and refinement stage. In the initial stage, MANO with 3D pose and shape priors is used to reconstruct physical plausible hand meshes, but it is challenging to regress accurate MANO parameters with severe hand-occlusion from RGB image. Then in the refinement stage, the model-free method guided by MANO is used to further improve the accuracy of the reconstruction. 

In the initial stage, we aim to obtain the high-quality MANO model in hand-object interaction scenario. In the regression of MANO parameters, the previous works \cite{boukhayma20193d}\cite{liu2021semi}\cite{park2022handoccnet} use CNN to encode the extracted hand image features and 2D heatmaps as a latent feature, and then transfer the latent feature to the pose decoder to obtain MANO parameters. However, those methods don’t distinguish the pose and shape parameters and the mapping from the abstract image feature to the MANO parameter space is highly non-linear. Some approaches \cite{gao20223d}\cite{huang2020hot}\cite{yu2023overcoming} employ MLP or Transformer to regress MANO parameters from concrete 3D joints, but those methods depend on the accurate 3D joints estimation in the previous stage. Due to the severe hand-occlusion and inherent depth ambiguity of monocular RGB images, it is not easy to obtain accurate 3D joints. In this work, we propose a novel MANO pose parameters regression module based on SemGCN that can achieve the mapping of MANO pose parameters directly from 2D joints. Previous methods \cite{huang2020hot}\cite{le2021sst}\cite{moon2020i2l}\cite{zhao2019semantic} have demonstrated that Graph Convolutional Network(GCN) can finish the encoding of 3D pose space features exploiting the topological structure information of 2D joints. Compared with ordinary GCN, SemGCN \cite{zhao2019semantic} can further adaptively model connection strength to capture semantic information among joints,  which strengthen the connection between occluded and unoccluded joints. We employ SemGCN to complete the encoding of 3D MANO pose parameters space through exploiting 2D joints information, which avoids the process of highly nonlinear mapping from abstract image feature and no longer depends on accurate 3D joints.

In the refinement stage, we aim to effectively refine the hand meshes and joints using model-free method guided by MANO from the initial stage. For enhancing the hand mesh and joint feature representation, the previous works \cite{ge20193d}\cite{tang2021towards}\cite{tse2022collaborative} employ GCN to aggregate adjacent mesh vertex and joint feature through modeling neighborhood vertex-vertex and joint-joint interactions respectively. However, those methods ignore the correlation between vertex and joint that weaken feature enhancement. To capture the correlation between vertices and joints, the common methods \cite{chen2021model}\cite{lin2021end}\cite{park2022handoccnet} use the pre-defined regression matrix from MANO, but the matrix is designed to only map MANO-derived meshes to joints. Recently, the methods \cite{jiang2023probabilistic}\cite{lin2021end} use 2D spatial attention mechanism to capture the correlation of vertex-joint to improve feature representation. Motivated by the above observation, we propose a vertex-joint mutual graph-attention model to jointly enhancing mesh vertex and joint feature from the initial stage. Specifically, our proposed refinement model is made up of a stack of basic blocks of the similar structure, which consists of a GCN layer and a mutual attention. GCN is employed to model the dependencies of vertex-vertex and joint-joint for aggregating intra-graph node features. And mutual attention  is devised to capture the correlation between vertices and joints for aggregating inter-graph node features.  

In summary, our contributions are as follows:
\begin{enumerate}
\itemsep=0pt
\item We propose a 3D hand reconstruction pipeline combining the benefits of model-based and model-free approaches for hand-object interaction scenario, which balance accuracy and physical plausibility to improve the precision of hand meshes and joints estimation.
\item We present a novel MANO pose parameters regression module from 2D joints directly, which avoids the process of highly nonlinear mapping from abstract image feature and no longer depends on accurate 3D joints.
\item We propose a vertex-joint mutual graph-attention model guided by MANO that can jointly refine hand mesh vertices and joints, which model the dependencies of vertex-vertex and joint-joint and capture the correlation of vertex-joint for aggregating intra-graph and inter-graph node features.
\item We demonstrate the effectiveness of our method by conducting comprehensive experiments on two benchmark datasets: Dex-YCB \cite{chao2021dexycb} and HO3Dv2 \cite{shivakumar2020honnotate} that contain severe hand occlusion. Compared with the state-of-the-art methods, our method achieves a competitive performance. Moreover, we further perform ablation studies to highlight the effectiveness of our proposed model and its major components on the HO3DV2 dataset.
\end{enumerate}  
\section{Related work}
\subsection{Model-based Methods} In recent years, many methods have been proposed to reconstruct the 3D hand by regressing the MANO parameters. Boukhayma et al. \cite{boukhayma20193d} firstly regress the MANO parameters in single-hand reconstruction via a deep neural network. Zhang et al. \cite{zhang2019end} propose an architecture of iterative regression module based on CNN to obtain the MANO parameters from 2D joint heat-maps. For hand-object interaction scenario, some approaches focus on modeling interactions between hands and objects. Liu et al. \cite{liu2021semi} devise a contextual reasoning module for modeling hand-object interaction. A most recent approach \cite{lin2023harmonious} propose a harmonious feature learning network that can extract hand and object harmonious features. Some approaches based on 2D spatial attention has achieved an impressive performance. Park et al. \cite{park2022handoccnet} introduce an feature injection mechanism that fully exploits the information at occluded regions to enhance image features. However, it is challenging to regress accurate MANO parameters that limited their reconstruction accuracy. Model-based approaches with pose and shape priors show great advantages for hand reconstruction. However, most previous methods focus on feature extraction and enhancement and ignore the highly nonlinear mapping process of MANO parameters. In this work, our proposed MANO pose parameters regression module from 2D joints directly is able to avoid the process of highly nonlinear mapping from abstract
image feature and no longer depend on accurate 3D joints.  

\subsection{Model-free Methods} To relax this heavy reliance on the MANO parameter space, some approaches directly fit the mesh vertices. Ge et al. \cite{ge20193d} and Choi et al. \cite{choi2020pose2mesh} use Graph-CNN to model neighborhood vertex-vertex interactions for regressing the hand meshes in a coarse-to-fine manner. Kulon et al. \cite{kulon2020weakly} apply spiral convolutions for neighbourhood selection. Tang et al. \cite{tang2021towards} utilize a small Graph-CNN to predict an offset mesh to quickly align the rough mesh with AR image space. Tse et al. \cite{tse2022collaborative} use GCN with attention-guided to iteratively refine results from dense hand and object graphs. But GCN can only capture the local interactions between neighboring vertices, Lin et al. \cite{lin2021end} use transformer to capture global interactions between the vertices and joints. Wang et al. \cite{wang2023interacting} introduce a dense mutual attention to model fine-grained dependencies between the hand and the object. Different from model methods, models-free methods can improve average accuracy of reconstruction but are usually weak to robust to occlusion and low resolution. In this work, our proposed vertex-joint mutual graph-attention model gudided by MANO are used to effectively refine 3D mesh vertices and joints.  

\subsection{GCN-based Methods} Graph convolution network (GCN) allow enhancing representations of the relationships between the nodes of graph-based data. GCN have been wildly used in 3D hand pose estimation and mesh reconstruction since hand mesh and joints naturally form a graph. Zhao et al. \cite{zhao2019semantic} proposed a SemGCN for capturing complex semantic relationships among human body joints for 3D human pose estimation. Doosti et al. \cite{doosti2020hope} proposed an adaptive Graph-UNet jointly estimating 3D hand joints and object bounding box corners from 2D coordinates. Huand et al. \cite{huang2020hot} encode initial 2D poses with GCNs for the following 3D reconstruction in Transformer. Most recently, Wang et al. \cite{wang2023handgcnformer} propose an node-offset graph convolutional cascadeing transformer and GCN, and devise a Topology-aware head based on SemGCN.

\begin{figure*}[t]
	\centering
		\includegraphics[width=\linewidth]{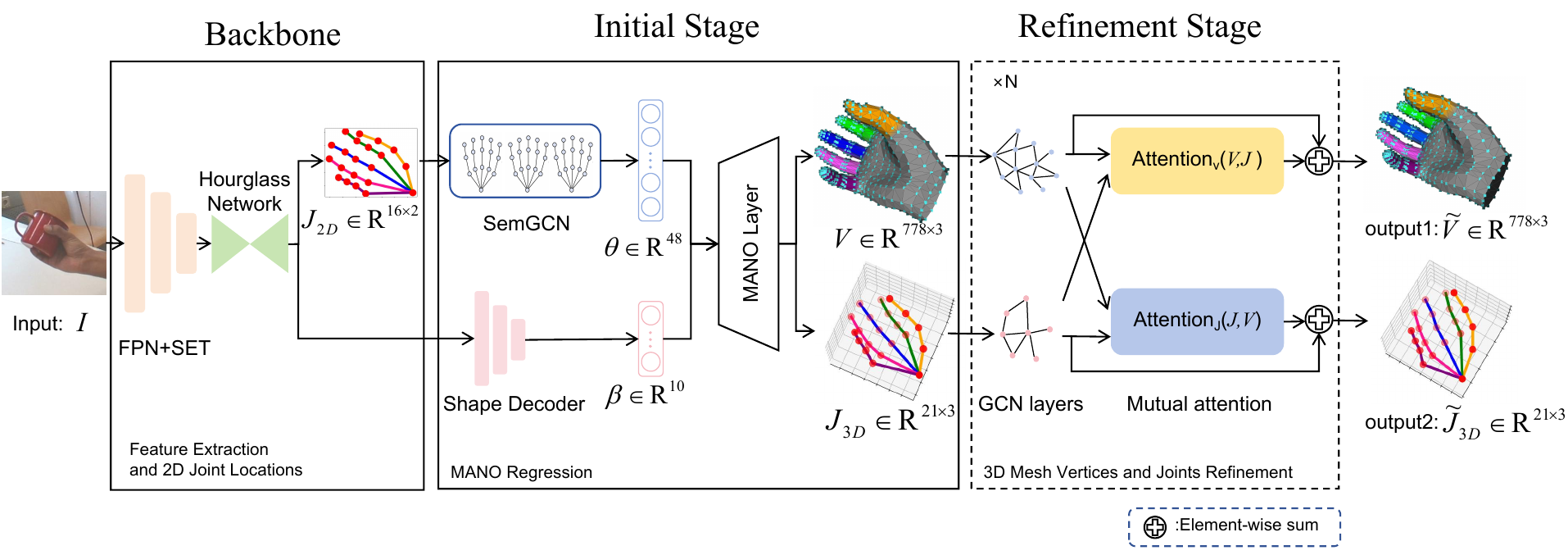}
	\caption{Overview of our method that consists of three stages. The Backbone extracts the hand feature $I$ and finish the 2D joint locations ($J_{2D}$) from the input single RGB image. At the initial stage, two separate branches are used to regress the MANO pose parameters and shape parameters respectively. Then, the rough 3D hand mesh vertices ($V$) and joints ($J_{3D}$) are obtained by forwarding the MANO parameters to MANO layer. In the refinement stage, we firstly construct mesh vertex and joint graphs according to the MANO model. In the refinement stage, we use a stack of basic block of the similar structure to generate the refinement hand meshes and joints, where a basic block consists of stack of GCN layers followed by the proposed mutual attention.}
	\label{FIG:1}
\end{figure*}
\section{Method}
In this section, we introduce the overall pipeline of our method as shown in \Cref{FIG:1}. Our model consists of three stages. At the backbone, we firstly extract image feature and obtain hand 2D joint locations from an input RGB image $I\in R^{H\times W\times 3}$ (Section 3.1). At the initial stage, we then regress the MANO model to obtain rough hand mesh vertices and 3D joints (Section 3.2). At the refinement stage,we jointly refine the mesh vertex and joint feature using the proposed vertex-joint mutual graph-attention model (Section 3.3). The final outputs of the refinement stage are the 3D vertex and joint coordinates. We train the proposed model end-to-end and show the details of overall loss function (Section 3.4).
\subsection{Backbone}
The backbone extracts feature $F$ and localize 2D joints $J_{2D}$ from a hand RGB image $I\in R^{256\times 256\times 3}$. We firstly feed the hand image to ResNet50 \cite{he2016deep} -based FPN \cite{lin2017feature}, which produces the feature map $F\in R^{32\times 32\times 256}$. After that, a SET \cite{park2022handoccnet} model is adopted to refine the feature $F$ by referencing the distant information  with self-attention mechanism. Then, a stacked Hourglass Network takes enhanced feature as input and output 2D heatmaps $H$ for each joint. At last, we use COM-based decoder \cite{zhang2020differentiable} to regress 2D joints coordinates $J_{2D}$ from 2D heatmaps $H$.
\subsection{Initial Stage}
\subsubsection{Preliminaries}
MANO as a low dimensional parametric hand model is devised to cope with low-resolution, occlusion, and noise, which is learned from around 1000 high-resolution 3D scans of hands of 31 subjects. The final 778 vertices and 16 joints of one hand are obtained by forwarding the pose parameters $\theta \in R^{48}$ and shape parameters $\beta \in R^{10}$ to MANO layer. And $\theta$ is joint rotations representation in axis-angle and $\beta$ is coefficients of a shape PCA. The MANO hand model is defined as follows: 
\begin{equation} \left\{\begin{aligned} 
&M(\beta ,\theta )=W(T_P(\beta ,\theta ),J(\beta),\theta,\omega ) \\  
  &T_P(\beta ,\theta)=\overline{T} +B_s(\beta)+B_p(\theta)
\end{aligned}\right.
\end{equation}
where $W$ is a linear blend skinning (LBS) function, $T_P$ corresponds to the articulated mesh template to blend, consisting of $K$ joints, $J$ represents the joint locations learned from the mesh vertices via a sparse linear regressor, and $\omega$ indicates the blend weights. $\overline{T}$ is a mean hand template, $B_s$and $B_p$ are shape blendshapes and pose blendshapes respectively.
\subsubsection{MANO parameters Regression}
\begin{figure}[t]
	\centering
		\includegraphics[width=\linewidth]{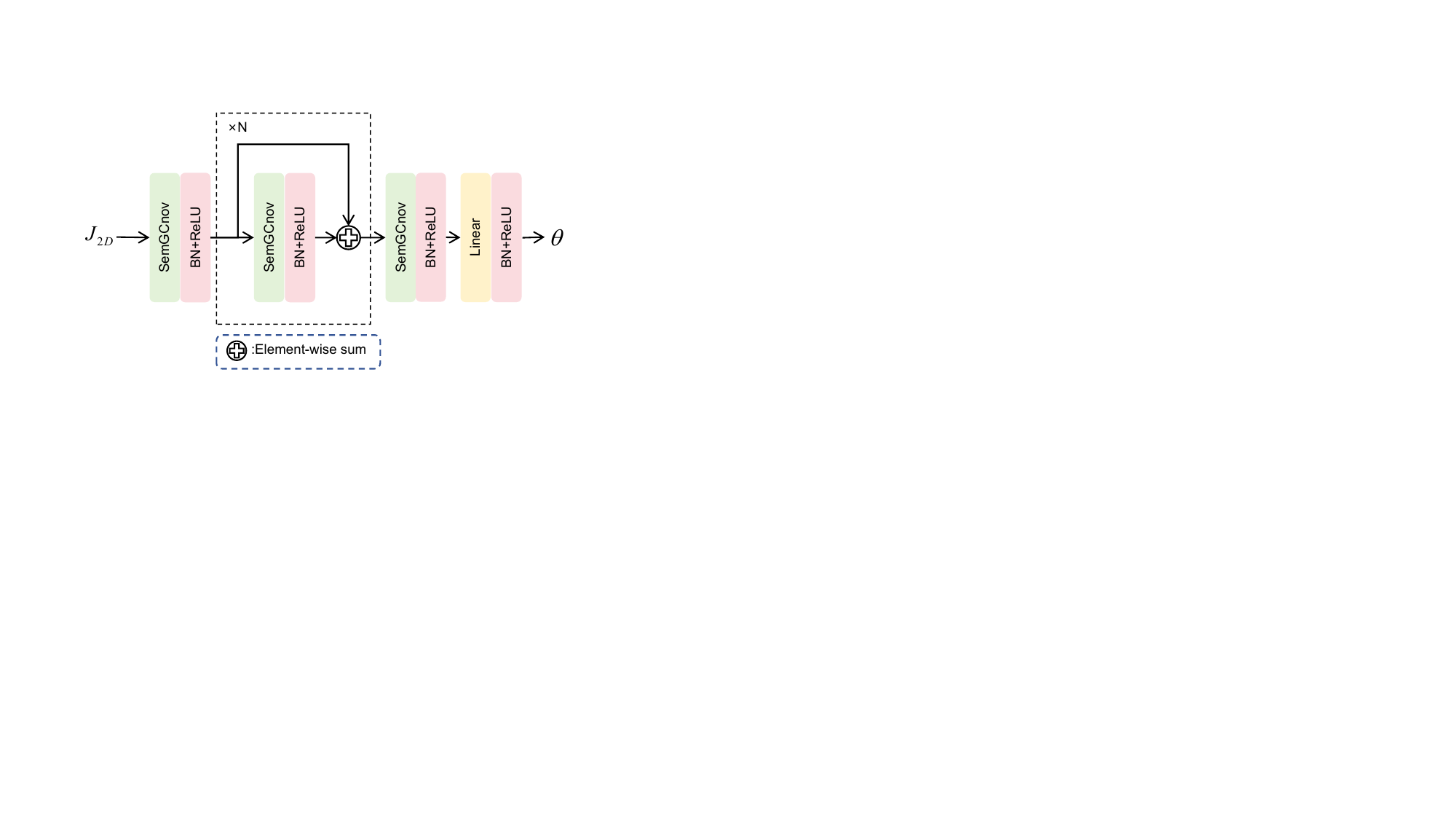}
	\caption{The SemGCN-based MANO pose parameters Regression module.}
	\label{FIG:2}
\end{figure}

It is challenging to directly regress MANO parameters from image feature, since the mapping from an image to the MANO parameters space is highly non-linear. In this work, we propose a novel MANO pose parameters regression module basded on SemGCN achieving the mapping of the MANO pose parameters from 2D joints directly. 2D joints can be considered as a kind of graph structure data, in which each joint can be regarded as a node. Node features contain rich location information and the neighboring nodes also provide useful features to estimate the relative offsets which can play a critical role for invisible joints. The topological hand joints structure plays an essential role in predicting accurate hand pose especially in severe occlusion. GCN naturally introduce the prior of hand kinematic topology and aggregates information about the nodes and their corresponding neighbor nodes under the guidance of topology. Compared with ordinary GCN, SemGConv \cite{zhao2019semantic} adds a learned weighting matrix $M$ to adaptively model connection strength between joints, which is written as:
\begin{equation} 
X^{(l+1)}=\sigma (\textbf{W}X^{(l)}\rho _i(\textbf{M}\otimes (\textbf{A}+\textbf{I}))
\end{equation}
where $\sigma$ is the activation function, $\textbf{W}$ is a transformation matrix, $\rho$ is the Softmax non-linearity which normalizes the weight of connections between a node and the neighboring nodes, $\otimes$ denotes element wise multiplication, $\textbf{A}$ is the adjacency matrix and $\textbf{I}$ is the identity matrix. We use SemGCN to encode the latent pose parameters space $\Theta \in R^{16\times 128}$ from 2D joints, which consists of ResGCN based on SemGconv. In hand-object interaction Scenario, SemGCN can capture semantic information about whether adjacent joints are occluded through utilizing the learned weighting matrix $\textbf{M}$, which strengthen connection between adjacently occluded and unoccluded joints. After that, we build a decoder to regress the pose parameters from the latent pose space, which consists of one SemGconv layer and MLP. The illustration of the module is shown in \Cref{FIG:2}.

For the shape parameters $\beta$ of MANO, we continue the method of previous works \cite{park2022handoccnet}\cite{liu2021semi}. The network with four residual blocks takes a concatenation of the enhanced hand feature $F$ and the 2D heatmaps $H$ as input and output 2048 dimensional vector, which is then passed to fully-connected layers to predict  $\beta$. Finally, the MANO model are obtained by forwarding the MANO parameters to MANO layer.
\subsection{Refinement Stage}
\begin{figure}[t]
	\centering
		\includegraphics[width=\linewidth]{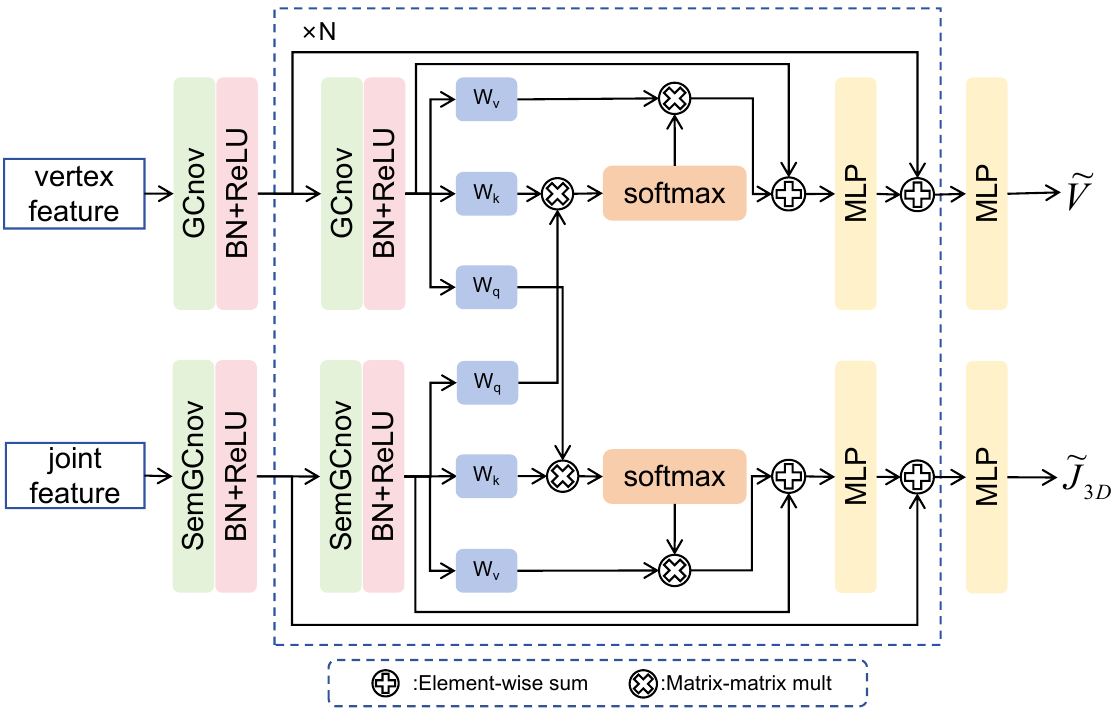}
	\caption{The MANO pose parameters Regression module based on SemGCN.}
	\label{FIG:3}
\end{figure}

Given the rough but physical plausible 3D hand mesh vertices $V\in R^{778\times 3}$ and joints $J_{3D} \in R^{21\times3}$ from the initial stage, we propose a vertex-joint mutual graph-attention model that can jointly refine hand mesh vertices and joints. To this end, we regard hand mesh and joints as two graphs and use GCN to capture the intra-graph dependencies. To further model the inter-graph interaction, we use mutual attention that allows feature aggregation between two graphs. The illustration of the model is shown in \Cref{FIG:3}.
\subsubsection{Graph Construction and GCN Layer}
The hand meshes and joints are modeled by separate graphs with vertices and joints as nodes, and their connections are defined in the MANO model mesh structure and joint topological structure. To construct the 3D vertex and 3D joint graph node features, we firstly extract global feature $F_{global}$ from the backbone FPN. Inspired of previous work\cite{jiang2023probabilistic}, we attach the initial MANO 3D coordinates of joints $J_{3D}$ and mesh vertices $V$ to the image-level global feature vector $F_{global}$ to obtain the new joint feature matrix $F_j\in R^{2051\times 21}$ and new vertex feature matrix$F_v\in R^{2051\times 778}$. After initializing the joint and vertex features, we then update the joint features $\hat{F} _j$ and vertex features  $\hat{F} _v$ via Gconv and SemGconv respectively. Intuitively, the graph convolutional layers can effectively model intra-graph dependencies with exploit neighboring correlation from the topology of the meshes and joints.
\subsubsection{Mutual Attention Layer}
Following one or several graph convolutional layers, we model vertex-joint interaction in the mutual attention layer. For each node from one graph, our mutual attention layer aims to aggregate features from the other graph via the attention mechanism. Specifically, for every node feature in the mesh graph, we first use three 1D convolutional layers to extract the query, key, and value, and collect all queries, keys, and values as $Q_v \in R^{778\times C}$, $K_v \in R^{778\times C}$ and  $V_v \in R^{778\times C}$ respectively, where each row of them is the query, key or value of a particular node. Similarly, we have the query, key and value for the joint graph as $Q_j \in R^{21\times C}$, $K_j \in R^{21\times C}$ and  $V_j \in R^{21\times C}$ respectively. We then compute the vertex-to-joint attention between the queries from the joint graph and the keys from the vertex graph following  as:
\begin{equation}
 A^{v\rightarrow j}= softmax(\frac{Q_j K_v^T}{\sqrt{C} } )   
\end{equation}
where $A^{v\rightarrow j}\in R^{21 \times 778}$ is the vertex-to-joint attention map, with the $i$-th row denoting the expected contribution proportion of all vertices nodes to the $i$-th joint node. We can then aggregate vertex node features weighted by the vertex-to-joint attention as:
\begin{equation}
  F^{v\rightarrow j}= A^{v\rightarrow j} V_v    
\end{equation}
where $F^{v\rightarrow j}\in R^{21\times C}$ is the aggregated features from the vertex graph. Similarly, we can compute the joint-to-vertex attention as:
\begin{equation}
 A^{j\rightarrow v}= softmax(\frac{Q_v K_j^T}{\sqrt{C} } )   
\end{equation}
where $A^{j\rightarrow v}\in R^{778 \times 21}$. And we can compute the joint-to-vertex aggregated feature as:
\begin{equation}
  F^{j\rightarrow v}= A^{j\rightarrow v} V_j    
\end{equation}
where $F^{j\rightarrow v}\in R^{778\times C}$. We finally fuse the aggregate feature with the original feature in each node as:
\begin{equation} \left\{\begin{aligned} 
&\widetilde{F}_j=f_j(\hat{F_j}\oplus F^{v \rightarrow j}) \\  
&\widetilde{F}_v=f_v(\hat{F_v}\oplus F^{j \rightarrow v})
\end{aligned}\right.
\end{equation}
where $\widetilde{F}_j$ and $\widetilde{F}_v$ are the refined joint and vertex feature, $f_j$ and $f_v$ are independent fusion units consisting of MLP. Finally, $\widetilde{F}_j$ and $\widetilde{F}_v$ are passed to MLP to predict the final 3D hand mesh vertices $\widetilde{V}\in R^{778\times 3}$ and joints $\widetilde{J}_{3D}\in R^{21\times 3}$ respectively.
\subsection{Overall Loss Function}
To effectively train the proposed model, the multi-task loss function employed here is defined as:
\begin{equation}
L=\lambda _1L_{MANO}+\lambda _2L_{r}+\lambda _3L_{e}+\lambda _4L_{n}
\end{equation}
where $L_{MANO}$ denote the MANO loss to supervise backone and initial stage, $L_{r}$ denote the refinement loss in the refinement stage, $L_{e}$ and $L_{n}$ denote the edge loss and normal loss that are used to penalize outlier vertices and irregular surfaces. Finally, $\lambda _1$, $\lambda _2$, $\lambda _3$ and $\lambda _4$ are coefficients that balance the weight of each loss function.
\subsubsection{MANO Loss}
We adopt an L2 loss to supervise the regression of 2D joint location and the MANO model, and the $L_{MANO}$ is defined as: 
\begin{equation}
 \begin{split}
L_{MANO}=&\left \| V-V^{gt} \right \|_2 +\left \| J_{3D}-J_{3D}^{gt} \right \|_2  
         +\left \| \theta -\theta ^{gt} \right \|_2  +\\&\left \| \beta-\beta ^{gt} \right \|_2+\left \| H-H ^{gt} \right \|_2+\left \| J_{2D}-J_{2D} ^{gt} \right \|_2 
\end{split}   
\end{equation}
where $V^{gt}$, $J_{3D}^{gt}$, $\theta ^{gt}$, $\beta ^{gt}$, $H ^{gt}$ and $J_{2D} ^{gt}$ represent the ground truth of mesh vertices, 3D joints, MANO pose parameters, shape parametesr,2D heatmaps and 2D joints respectively.
\subsubsection{Refinement Loss}
The final outputs in the refinement stage are the coordinates of 3D joint $\widetilde{J}_{3D}$ and the hand mesh vertex $\widetilde{V}$ . We also adopt an L2 loss to supervise the refinement stage and the $L_r$ is defined as: 
\begin{equation}
L_r=\left \|\widetilde{V}-V^{gt}\right \|_2 + \left \|\widetilde{J}_{3D}-J_{3D}^{gt}\right \|_2
\end{equation}
\subsubsection{Edge Loss and Normal Loss}
We employ the edge loss $L_e$ and the normal loss $L_n$ to penalize flying vertices and irregular surfaces following \cite{tang2021towards}, which is defined as :
\begin{equation}
L_e=\left \| \left | \widetilde{e}_i\right |-\left | e^{gt}_i\right |  \right \|_1 
\end{equation}
\begin{equation}
L_n=\left \|\left \langle \widetilde{e}_i,n^{gt}_i  \right \rangle  \right \|_1 
\end{equation}
where $\widetilde{e}_i$ denote the $i$-th mesh edge vector of the refined hand meshes, $e^{gt}_i$ and $n^{gt}_i$ are the ground truth of the edge vector and the normal of the corresponding edge respectively.
\section{Experiments}
In this section, we firstly introduce the benchmark datasets for 3D hand reconstruction to train our proposed model and define the evaluation metrics. Then we present the implementation details and the quantitative and  qualitative results of comparing with state-of-the-art methods. Finally, we present the results of ablation studies on each component to demonstrate the effectiveness of our proposed modules.
\subsection{Datasets}
The following recently hand reconstruction benchmark datasets are exploited to train and evaluate our proposed model: HO3DV2 and Dex-YCB, which contain significant occlusions in hand-object interaction scenario. The HO3DV2 is a challenging dataset of hand-object interaction,which consists of 77K images from 68 video sequences and are split into 66,034 images for training and 11,524 images for testing. And we submit the test results to the official website to report performance. Dex-YCB is a latest large scale dataset that includes 582K images from over 1000 video sequences that covers 10 subjects and 20 objects. And we present the results according to the official s0 splitting protocol. Specifically, Dex-YCB is a recently large scale dataset and our model is pre-trained on Dex-YCB for a few epochs to alleviate the overfitting on HO3DV2 following \cite{xu2023h2onet}.
\subsection{Evaluation Metrics}
In order to evaluate our proposed model and consistently compare the results with state-of-the-art methods, we adopt the evaluation metrics on each benchmark dataset that are commonly used for 3D hand reconstruction and majorly reported by related works\cite{park2022handoccnet}\cite{jiang2023probabilistic}. These are the mean per joint/vertex position error(J-PE/V-PE), mean per joint/vertex position error after procrustes alignment(PA-J-PE/PA-V-PE) and vertices at distance thresholds of 5mm and 15mm after procrustes alignment (PA-F@5 and PA-F@15), where the position error is the mean 3D Euclidean distance between the predicted coordinates and the groundtruth coordinates, and procrustes alignment is employed to ignor the global rotation and scale differences.

\begin{table*}[width= \textwidth,cols=4,pos=t]
\caption{Hand reconstruction results on the HO3Dv2 dataset compared with SOTA methods. -: the results are unavailable from previous papers. We mark the best results in bold for better comparison.}\label{tbl1}

\begin{tabular*}{\linewidth}{LLLLLLLL }
\toprule
Method            & Category      &J-PE & PA-J-PE & V-PE &PA-V-PE &PA-F@5 & PA-F@15\\
\midrule
ArtiBoost \cite{yang2022artiboost}     & Model-based   & -    & 11.4   &-     & 10.9   &0.488  &0.944\\
Hasson et al. \cite{hasson2019learning} & Model-based   & -    & 11.0   &-     & 11.2   &0.464  & 0.939\\
Hampali et al. \cite{shivakumar2020honnotate}& Model-based   & -    & 10.7   &-     & 10.6   &0.506  &0.942\\
I2UV-HandNet \cite{chen2021i2uv}  & Model-based   & -    & 9.9   &-     & 10.1    &0.500  & 0.943\\
Liu et al. \cite{liu2021semi}    & Model-based   & 30.0   & 9.9   &28.9     & 9.5   &0.528  &0.956\\
HandOccNet \cite{park2022handoccnet}     & Model-based   & 24.9 & 9.1   &24.2     & 8.8   &0.564  & 0.963\\
HFL-Net \cite{lin2023harmonious}        & Model-based   & -    & 8.9   &-     & 8.7       &0.575  & 0.965\\
Pose2Mesh \cite{choi2020pose2mesh}       & Model-free   & -    & 12.5   &-     & 12.7   &0.441  &0.909\\
I2L-MeshNet \cite{moon2020i2l}  & Model-free    & 26.8  & 11.2   &-     & 13.9    &0.409  & 0.932\\
Keypoint Trans \cite{hampali2022keypoint}  & Model-free & -    & 10.8      &-     & -       &-       & -\\
METRO \cite{lin2021end}          & Model-free    & -  & 10.4   &-     & 11.1    &0.484  & 0.946\\
H2ONet \cite{xu2023h2onet}          & Model-free    & 23.0  & 9.0   &22.4     & 9.0    &0.554  & 0.960\\
Jiang et al. \cite{jiang2023probabilistic}          & combined    & \textbf{19.2}  & \textbf{8.3}   &1\textbf{8.3}     & \textbf{8.2}    &\textbf{0.608}  & 0.965\\
\textbf{Ours}                &combined     & 19.7  & 8.8   &19.4     & 8.7    &0.555  & \textbf{0.970}\\
\bottomrule
\end{tabular*}
\end{table*}

\begin{table*}[width= \textwidth,cols=4,pos=t]
\caption{Results on the Dex-YCB dataset compared with SOTA methods. -: the results are unavailable from previous papers. We mark the best results in bold for better comparison.}\label{tbl2}

\begin{tabular*}{\linewidth}{LLLLLL }
\toprule
Method            & Category                             & V-PE &PA-V-PE &J-PE & PA-J-PE\\
\midrule
Spurr et al. \cite{spurr2020weakly}     & Model-based   &-     & -  &17.3 & 6.8 \\
Liu et al. \cite{liu2021semi}        & Model-based       &-     & -   & 15.3    & 6.6 \\
HandOccNet \cite{park2022handoccnet}    & Model-based   &13.1     & 5.5  & 14.0    & 5.8 \\
HFL-Net \cite{lin2023harmonious}         & Model-based    &-     & -  & \textbf{12.7}   & \textbf{5.5} \\
METRO \cite{lin2021end}            & Model-free           &-     & -    & 15.2   & 6.7\\
MobRecon \cite{chen2022mobrecon}         & Model-free      & 13.1    & 5.6 & 14.2    & 6.4 \\
H2ONet \cite{xu2023h2onet}          & Model-free         & 13.0    & 5.5  & 14.0    & 5.7\\
\textbf{Ours}                     & combined               & \textbf{12.4}    & \textbf{5.4}  & 13.2    & 5.6  \\
\bottomrule
\end{tabular*}
\end{table*}

\subsection{Implementation Details}
We implement our method in PyTorch. All experiments are run on NVIDIA RTX 2080Ti GPU and Inter xeon gold 6226 CPU.We train the network using the Adam \cite{2014Adam} optimizer and set the batch size as 24. The initial learning rate is set as 1e-4. For Dex-YCB, we train the model in 50 epochs and the learning rate decay by 0.9 after every 10 epochs. For HO3DV2, we firstly pre-train the model on Dex-YCB for 20 epochs. Then we train the model in 100 epochs and the learning rate decay by 0.7 after every 10 epochs. To makes hands in the center of the input image, we use the hand ground truth 2D joints to crop each sample image and resize it to 256×256. And common data augmentation strategies are adopted including random scaling([0.9,1.1], rotation([-180,180]), translation([-10,10]), and color jittering. 
\subsection{Comparison with State-of-the-art Methods}  
To evaluate the hand reconstruction quality from a single RGB image, we compare our method with the state-of-the-art methods quantitatively on the HO3DV2 and Dex-YCB. As the ground truths of HO3DV2 test set are not publicly available, we obtain the test results from the official evaluation server. Besides, we follow \cite{xu2023h2onet} to pre-train the model on Dex-YCB for a few epochs to alleviate the overfitting on HO3DV2. We report the performance before and after the procrustes alignment(PA) to better show our method effectiveness. The experimental results on HO3DV2 and Dex-YCB are shown in Table 1 and Table 2 respectively. 

As shown in Table1, comparing with all state-of-art only model-based and model-free methods, our proposed network cascading MANO and Model-free method achieve the best performance on majority evaluation metrics on HO3DV2, and only second to the most latest Jiang et al. \cite{jiang2023probabilistic} that propose a probabilistic model that estimate the prior probability distribution of joints and vertices with incorporating a model-based network as a prior-net to. Especially on mean per joint/vertex position error without procrustes alignment metrics(J-PE/V-PE), we have made considerable progress compared with most previous methods, and we are slightly better than Jiang et al. \cite{jiang2023probabilistic} on the PA-F@15. As shown in Table2, our method also achieve competitive performance on Dex-YCB and only second to the most latest HFL-Net \cite{lin2023harmonious}. We attribute the advantage of our proposed network to combine the benefits of model-based and model-free approaches, which MANO is used to reconstruct coarse but physical plausible hand meshes and pose in the initial stage and model-free method guided by MANO is used to further improve the accuracy of the reconstruction in the refinement stage.
\begin{figure*}[t]
	\centering
		\includegraphics[width=\linewidth]{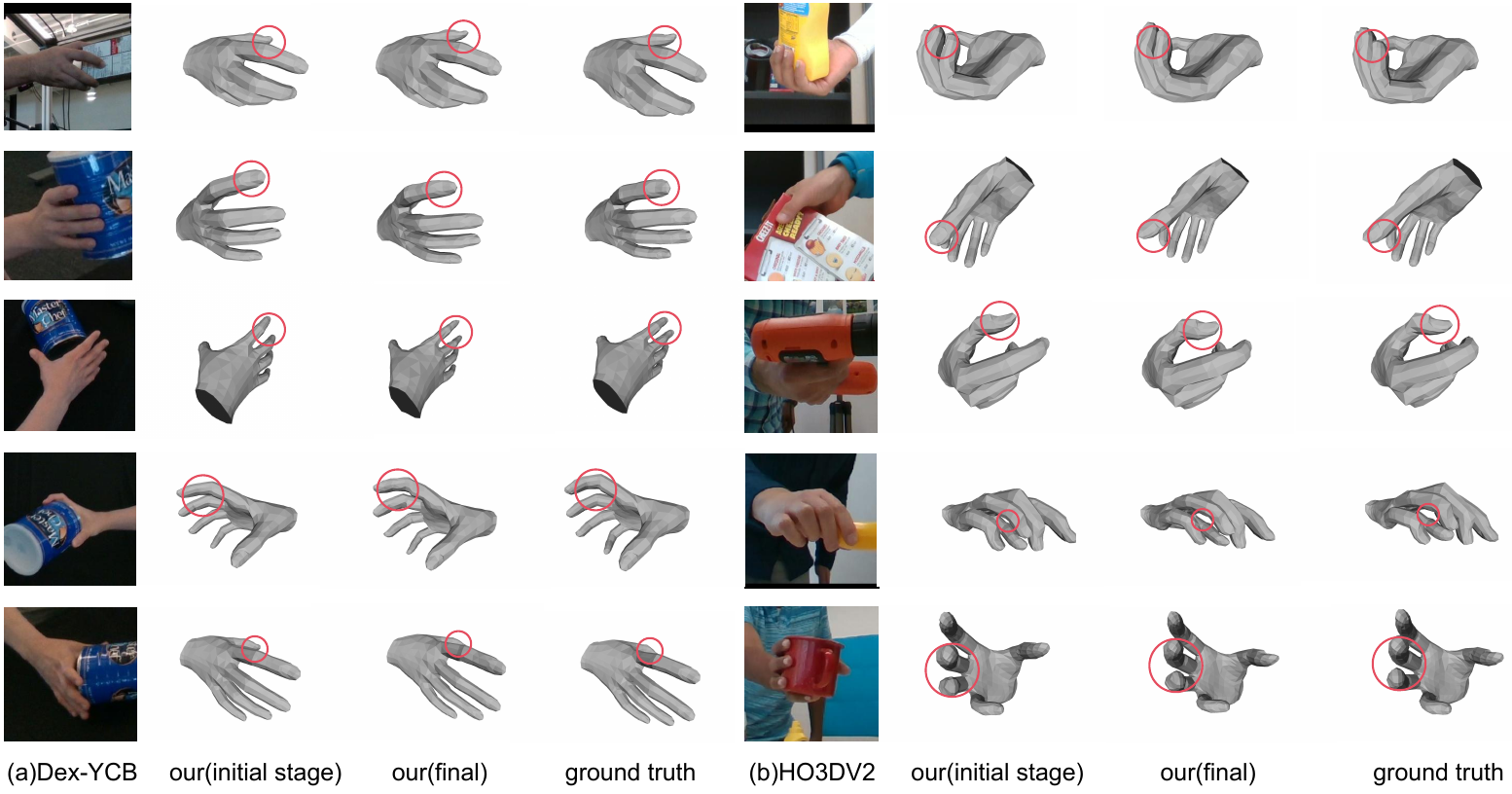}
	\caption{Qualitative results of our method on the HO3D and Dex-YCB datasets.}
	\label{FIG:4}
\end{figure*}
\newcommand{\RNum}[1]{\uppercase\expandafter{\romannumeral #1\relax}}

\begin{table}[width= \linewidth,cols=4,pos=t]
\caption{Analysis of the MANO pose parameters regression module.}\label{tbl3}
\begin{tabular*}{\linewidth}{LLLL }
\toprule
Exp            & Set      &PA-J-PE & PA-V-PE  \\
\midrule
{\romannumeral1})            & B   & 10.4    & 10.0    \\
{\romannumeral2})            & B+OG   & 9.7    & 9.7   \\
{\romannumeral3})            & B+SG   & 9.5    & 9.4    \\
{\romannumeral4})           & B+SG+G   & 9.3    & 9.4    \\
{\romannumeral5})           & B+SG+G+MA   & 9.0    & 9.1   \\
{\romannumeral6})            & B+SG+G+MA+P   & \textbf{8.8}    & \textbf{8.7}  \\
\bottomrule
\end{tabular*}
\end{table}

\subsection{Ablation Studies}
\begin{figure}[t]
	\centering
		\includegraphics[width=\linewidth]{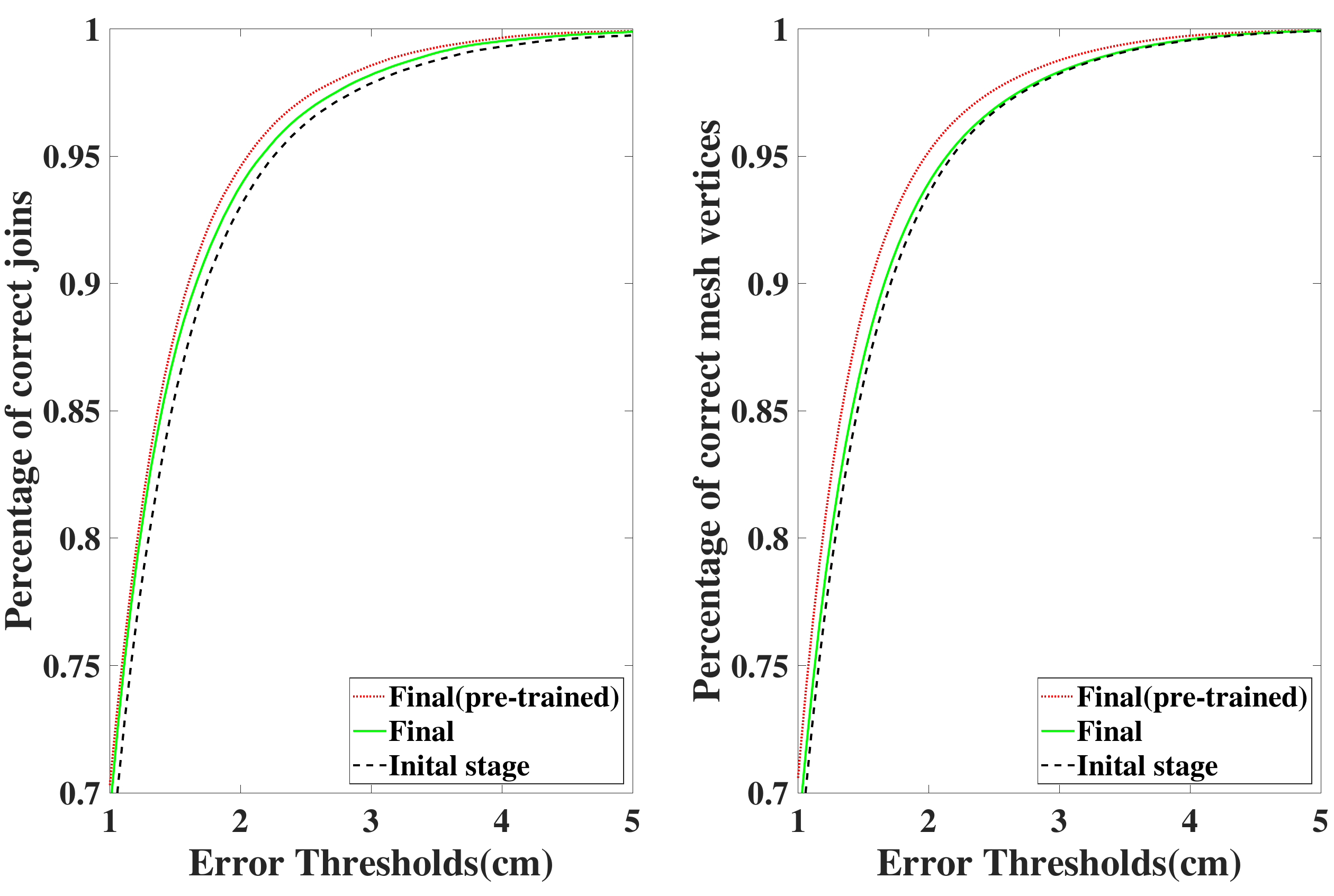}
	\caption{3D joints and mesh vertices PCK after procrustes aligned of our method on HO3DV2 dataset.}
	\label{FIG:5}
\end{figure}

We perform ablation studies on HO3DV2 to study the effectiveness of our proposed model and its major components. As \Cref{tbl3} shows, we denote Baseline (B) as our model after removing the following major components: SemGCN(SG) in MANO pose parameters regression module,GCN layer(G) and mutual attention layer(MA). Besides, OG denotes the ordinary GCN is used in MANO pose parameters regression module and P denotes that the model is pre-trained on Dex-YCB for a few epochs. Besides, we also provide 3D joints and mesh vertices PCK after procrustes aligned of our method on HO3DV2 dataset in \Cref{FIG:5}. 
\subsubsection{Effectiveness of MANO Pose Parameters regression Module}
We follow the previous methods \cite{liu2021semi}\cite{park2022handoccnet} to regression MANO pose and shape parameters on Baseline(B), which use CNN to encode the MANO parameters space from the concatenation of hand feature and 2D heatmaps. Comparing the first three rows in Table 3, we can see that GCN can more effectively encode MANO pose parameters space from 2D joint with alleviating the process of highly nonlinear mapping than Baseline and SemGCN achieve the best performance. SemGCN adds a learned weighting matrix M to adaptively model connection strength between adjacent joints, that is more suitable for our regression task comparing with ordinary GCN.
\begin{figure*}[t]
	\centering
		\includegraphics[width=\linewidth]{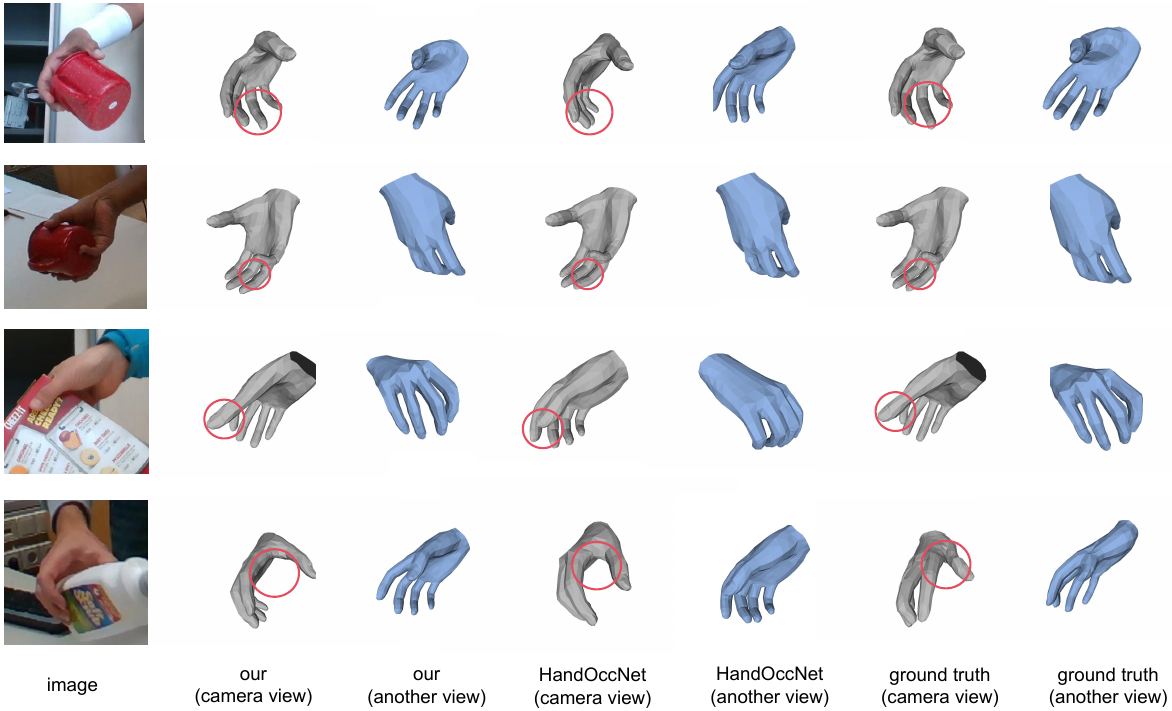}
	\caption{Qualitative comparison of with state-of-art method on HO3DV2 dataset.}
	\label{FIG:6}
\end{figure*}
\subsubsection{Effectiveness of Vertex-Joint Mutual Graph-Attention Module}
Our proposed vertex-joint mutual graph-attention module contists two main components: GCN layer (G) and mutual attention (MA). Comparing Rows {\romannumeral3}), {\romannumeral4}) and {\romannumeral5}) in \Cref{tbl3}, we can see that only GCN layer with modeling the dependencies of vertex-vertex and joint-joint can improve the precision of joints from the initial stage. When the mutual attention  (MA) is included, we observe the best performing result that demonstrate our proposed mutual graph-attention layer can effectively further aggregate inter-graph node features through capture the correlation between the vertices and joints. Besides, we follow \cite{xu2023h2onet} to pre-train the model on Dex-YCB for a few epochs to alleviate the overfitting on HO3DV2 and the comparison result of Rows {\romannumeral5}) and {\romannumeral6}) show its necessity.

\subsection{Qualitative Results}
To further evaluate the hand mesh reconstruction quality, we visualize 3D hand meshes on camera view and another view, and we show the qualitative results on HO3DV2 and Dex-YCB in \Cref{FIG:4} and \Cref{FIG:6} respectively. \Cref{FIG:4} show the qualitative comparison results of the initial stage and final output. And the results of initial stage show that our proposed MANO regression module can provide the initial MANO model that robust to object-occlusion for hand-object interaction scenario.Then the results of final output show that our proposed vertex-joint mutual graph-attention model guided by MANO can effectively refine hand mesh vertices and joints jointly. Moreover, \Cref{FIG:6} show the qualitative comparison results between our method and state-of-art HandOccNet on HO3DV2. It can be seen from the figure that our method makes more accurate hand meshes estimation than HandOccNet on camera view. 
\section{Conclusion}
In this paper, we propose a 3D hand reconstruction pipeline combining the benefits of model-based and model-free approaches for hand-object interaction scenario. Our proposed MANO pose parameters regression module based on SemGCN can effectively encode MANO pose parameters space from 2D joint directly and achieve better performance than the common methods using CNN to regress MANO pose parameters from image feature. Moreover, we further propose a vertex-joint mutual graph-attention model guided by MANO that can jointly refine hand mesh vertices and joints, which model the dependencies of vertex-vertex and joint-joint and capture the correlation between the vertex-joint. Compared our method with the state-of-the-art methods on recently benchmark HO3DV2 and Dex-YCB datasets, our method achieves competitive performance. On HO3Dv2, our method outperforms all only model-base approaches and model-free approaches and achieves second performance on majorly metrics, especially we have made considerable progress on J-PE/V-PE. On Dex-YCB, our method also achieve best performance on V-PE/PA-V-PE and second best performance on J-PE/PA-J-PE. Moreover, ablation studies on HO3DV2 demonstrate the effectiveness of our proposed model and its major components. Besides, the qualitative results on camera view and another view further show the hand reconstruction quality of our method in challenging environments.
\section*{Acknowledgments}
This work was supported in part by the Natural Science Foundation of Guangxi (Grant No. 2022JJB170009), and in part by the funding of basic ability promotion project for young and middle-aged teachers in Guangxi's colleges and universities (Grant No. 2022KY0008). Feng also acknowledges support by the Bagui Scholars Program of Guangxi Zhuang Autonomous Region.




\bibliographystyle{cas-model2-names}


\end{document}